\documentclass[10pt,twocolumn,letterpaper]{article}

\usepackage{cvpr} 
\usepackage{times}
\usepackage{multirow}
\usepackage{epsfig}
\usepackage{graphicx}
\usepackage{amsmath}
\usepackage{amssymb}
\usepackage[table]{xcolor}
\usepackage{colortbl}
\usepackage{float}
\usepackage{url}
\usepackage{hyperref}
\hypersetup{
    colorlinks=true,
    linkcolor=blue,
    filecolor=magenta,      
    urlcolor=cyan,
    pdftitle={Overleaf Example},
    pdfpagemode=FullScreen,
    }
\newcommand{\mycc}{\cellcolor{lightgray}}

\begin{document}

\title{Can Foundation Models Predict Fitness for Duty?}

\author{Juan E. Tapia, and Christoph Busch\\
da/sec-Biometrics and Internet Security Research Group, Darmstadt, Germany\\
Hochschule Darmstadt\\
{\tt\small juan.tapia-farias@h-da.de}
}

\maketitle
\thispagestyle{empty}

\begin{abstract}
Biometric capture devices have been utilised to estimate a person's alertness through near-infrared iris images, expanding their use beyond just biometric recognition. However, capturing a substantial number of corresponding images related to alcohol consumption, drug use, and sleep deprivation to create a dataset for training an AI model presents a significant challenge. Typically, a large quantity of images is required to effectively implement a deep learning approach. Currently, training downstream models with a huge number of images based on foundational models provides a real opportunity to enhance this area, thanks to the generalisation capabilities of self-supervised models. This work examines the application of deep learning and foundational models in predicting fitness for duty, which is defined as the subject condition related to determining the alertness for work.
\end{abstract}

\section{Introduction}
\label{sec:intro}
Alcohol or drug consumption and sleep deprivation have been raised across the world, affecting the environment, especially the families and work areas. In labour environments, this consumption can produce severe accidents and take lives, especially in specific activities such as mining, logistics, commercial aviation and others where any alertness reduction may be severe~\footnote{\url{https://www.mining-technology.com/features/fatigue-kills-people-all-the-time-but-monitoring-is-making-strides/}}.

Drugs and alcohol are consumed widely in order to recover alertness and reduce the sleepiness condition that accumulates over several labour days and shifts, as many people indicate in several surveys or drug-free programs \footnote{\url{https://www.samhsa.gov/substance-use/drug-free-workplace}}. 

Fatigue, drowsiness, and sleepiness account for approximately~20\% of all injury-related crashes, according to the Road Safety Annual Report 2023 published by the Organisation for Economic Co-operation and Development (OECD)~\footnote{\url{https://www.itf-oecd.org/sites/default/files/docs/irtad-road-safety-annual-report-2023.pdf}}. The likelihood of fatal crashes attributed to alertness reduction can be as high as 30\%. 

The fatigue detection oriented to identify the physical and emotional condition of the workers has been raised in the last years to protect one of the most valuable assets in any company: "The human life". This area is named Fitness For Duty (FFD) and identifies a "FIT" person as a subject in physical condition to work and a "Unfit" person as a subject with an alertness reduction state because of some external factors such as alcohol, drugs and others that change the central nervous system. 

Traditionally, companies have used methods based on puzzles, math operations, wristbands, and capture devices on trucks, cars, helmets, or smart watches for fitness analysis. Nevertheless, these systems can not identify "who" is performing the action because they cannot capture any biometric data. Also, this device started to measure the signal of interest after beginning the labour (Reactively).  

In the last few years, some researchers have explored new approaches based on FFD topics, including biometric observation of body traits such as face and iris ~\cite{CAUSA2024122808, Zurita, TAPIA2025126511}. The iris has been prevalent because, in the capture process, one does not need to remove protective devices, such as dust masks or safety goggles, to measure the periocular area. This approach allows us to observe a captured subject and predict the fitness conditions declared as "Fit" or "Unfit" before starting the labour days (Pro-active), which is measured based on traditional NIR eyes samples. The new FFD measures the changes in the eyes based on the influence of external factors in the central nervous system (CNS). This action is initiated automatically by an external factor such as light or alcohol, drug consumption, etc.

On the other hand, one of the current challenges is obtaining a large number of images for training Convolutional Neural Networks (CNNs) based on deep learning or vision transformers \cite{resnet, swin}. Collecting images for the purpose of developing FFD is even more difficult because it is not trivial to get the collaboration of volunteers under the influence of alcohol or drugs for the training process. 

In FFD, the dataset contains control group images (FIT) that belong to one subject and have no relationship with other samples representing classes such as alcohol, drug, or sleepiness that belong to different subjects, and because of that, it is challenging to predict the correct class. This limitation makes the foundation model trained in the self-supervised method, plus fine-tuning techniques, an excellent candidate for exploring and improving this limitation.

This work focuses on studying FFD based on NIR eyes images and how the foundation model can improve the results of this algorithm based on the lack of images in this area, which a self-supervised algorithm can solve.

We propose a fine-tuning approach that uses DinoV2 and the OpenClip downstream model as feature extractors in which all the layers are frozen and fine-tune this general-purpose embedding extracted into our FFD domain using a small neural network head with three layers. The fine-tuning makes sense because of the lack of images available.

The remainder of this article is organised as follows: Section~\ref{sec:related} explores the state of the art. The proposed method for FFD detection is presented in Section~\ref{sec:method}. The experimental results are discussed in Section ~\ref{sec:experiment}. Conclusions and remarks are given in Section~\ref{sec:conclusion}.
 
\section{Related Work}
\label{sec:related}
In this section, we expand on the most critical contributions from the foundation models to feature extraction with a combination of CNNs and the state-of-the-art in FFD.

Researchers introduced the term "foundation model" to describe machine learning models that are trained on a wide range of generalised and unlabelled data \cite{FM-survey, iris-sam, iris_chatgpt}. These models are capable of performing various general tasks, including understanding language, generating text and images, and engaging in natural language conversations. 

The term "fine-tuning" in deep learning is a type of transfer learning. It involves taking a pre-trained model, which has already been trained on a large dataset for a general task like image recognition or natural language understanding, and making minor adjustments to its internal parameters. The goal is to enhance the model’s performance on a new, related task without having to start the training process from scratch \cite{FM-survey, chen2023vlp}.

\subsection{Fitness for Duty}

Researchers have explored the analysis of methods based on single NIR eyes images or time sequences measuring pupil changes per frame at low rates \cite{CAUSA2024122808, Zurita, TAPIA2025126511}. Other works have focused directly on pupillometry based on saccade distances, velocity, amplitude and other methods in order to obtain the best predictions of images of the present class through the use of Convolutional Neural Networks, Recurrent Neural Networks (RNN), Long-Short-Term-Memory-Networks, and combination of them \cite{Czajka2015, HollingsworthBowyerFlynn2009}. 

Alcohol, drug and sleepiness consumption have been explored by some researchers focusing on iris recognition~\cite{Arora, Tomeo-ReyesRossChandran2016, NIDA2021}, for alcohol classification~\cite{Pinheiro, tapia2022alcohol} and sleepiness \cite{FranzenBuysseDahlEtAl2009}.

Zurita et al.~\cite{Zurita} proposed an automated algorithm that extracts spatial and temporal features from a stream of NIR iris using Convolutional Neural Networks (CNN) in combination with Long-Short-Term-Memory-Networks~(LSTM) to estimate FFD.

Causa et al.~\cite{CAUSA2024122808} proposed a method that uses behavioural curves derived from NIR eyes images to address the limitations of alertness assessment. This approach enables the integration of a Fatigue Detection (FD) system into a biometrics device using a stream of images per subject. This method analyses the time series based on machine learning algorithms.

Tapia et al.~\cite{TAPIA2025126511} recently proposed an FFD method based on MobileNetV2 with several feature-extracted filter sizes per channel to classify alcohol, drug and sleepiness images using only one eye NIR image per subject. The model can estimate subjects’ proactive alertness before they begin their work duties while also verifying their identity. 

\subsection{Foundation models}

In order to analyse the impact of the foundation model in downstream tasks such as fitness for duty in self-supervised scenarios, we are exploring and analysing the power prediction of DinoV2 and OpenClip on FFD. Both are state-of-the-art methods.

The DinoV2 foundation model~\cite{dinov2} generates universal features that are suitable for various visual tasks. These tasks include image-level applications such as image classification and video understanding, as well as pixel-level tasks like depth estimation and semantic segmentation. Notably, DinoV2 was trained on 140 million unlabeled images.

The Contrastive Language-Image Pre-Training (CLIP) model~\cite{Radford2021LearningTV} learns visual concepts through natural language supervision, allowing it to be applied to any visual classification task by simply providing the names of the categories to be recognised. The CLIP model was trained with 400 million pairs of images and text.

Recent studies have begun to explore the potential of foundation models in biometrics, but there is still significant room for improvement in their applications~\cite{CHETTAOUI, arc2face}. 

The Iris-SAM method was proposed to segment iris images based on foundation models~\cite{iris-sam}. This model has been fine-tuned on ocular images for iris segmentation and has achieved a promising segmentation accuracy that is a challenge for the state of the art on the ND-IRIS-0405 dataset. 

Farmanifard et al.~\cite{iris_chatgpt} proposed an extensive evaluation based on Chat-GPT-4 ability to compare and analyse iris patterns and compare its performance against established iris recognition systems. This assessment aimed to determine the potential use of iris biometric recognition.

Very recently, Chettaoui et al.~\cite{CHETTAOUI} proposed using a foundation model for the Face recognition task, showing that it is possible to adapt this huge model to biometric modalities.

Building on these foundational approaches, this work highlights and evaluates the potential of foundation models across different tasks. For the first time, it investigates the effects of fine-tuning classification specifically for FFD, demonstrating the potential for improvement in results.

In summary, the main contributions of this work are:

Demonstrate the effectiveness of the foundation model-based framework based on Zero-shot and fine-tuning on an unrelated top-down task by adapting only a minimal number of parameters during the training phase. The results show that the framework's performance in fine-tuning for FFD improved significantly by simply adding a small head to various foundation models.

\begin{figure*}[]
      \centering
      \includegraphics[scale=0.64]{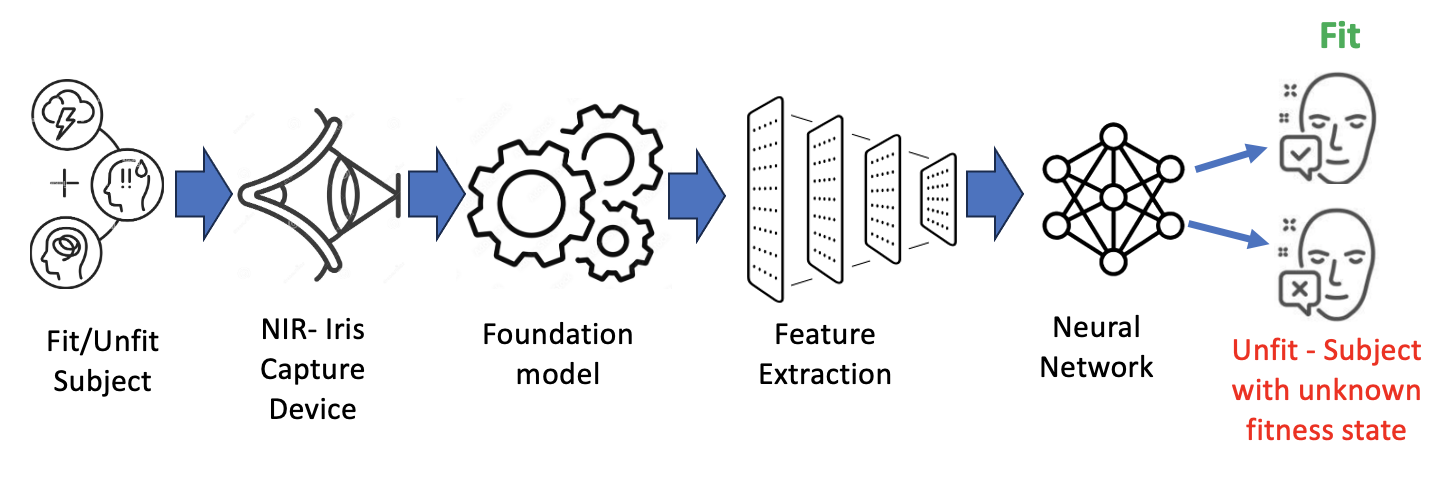}
      \caption{FFD workflow using foundation models.}
      \label{fig:workflow}
\end{figure*}

\section{Method}
\label{sec:method}

This work proposed a method to evaluate the generalisation capabilities of DinoV2 and OpenClip on the FFD approach. For this task, we used different versions of these downstream models to extract features from the control group subjects and subjects under the influence of alcohol, drugs and sleepiness. All the layers of DinoV2 and OpenClip were frozen and delivered an output of $1\times768$. Those embeddings were used for finetuning in a small neural network with three layers, exploring different parameters. For this exploration, three experiments were defined to measure the generalisation capabilities by comparing traditional CNN pre-trained based on ImageNet weights for the initialisation training process. The pipeline for the entire process is shown in Figure~\ref{fig:workflow}.

\subsection{Dataset}
\label{sec:dataset}

For this research, we utilized a dataset known as the “FFD NIR Iris Images Stream Database”~(FFD-NIR-Stream), which consists of 10-second streaming sequences of periocular NIR images for left and right eyes \cite{TAPIA2025126511}. All the images were captured taking into account the most similar conditions for a labour day. Considering at least an adaptation time for pupil size of 10 seconds to normalise the eyes for external conditions such as sunny days or sunglasses. 
However, from this dataset, we selected only the best image for each side (left and right)~\cite{TAPIA2025126511} based on the Laplacian of Gaussian operator as defined in ISO/IEC ISO/IEC 29794-6. In total, we have 62,995 images for training, 8,923 for validation and 72,093 for testing.

The captured NIR image frames are focused on the periocular area (eye mask) but represent each eye separately, including the pupils, irises, sclera and the skin that surrounds the eye area. The skin and pupil opening sizes are also relevant according to our observation because people under the influence of alcohol and drugs tend to close their eyes more than control subjects.
The capture device saves the best image automatically for the left and right eyes, which is determined using the LOG as previously mentioned.

\begin{figure}[H]
\centering
\includegraphics[scale=0.16]{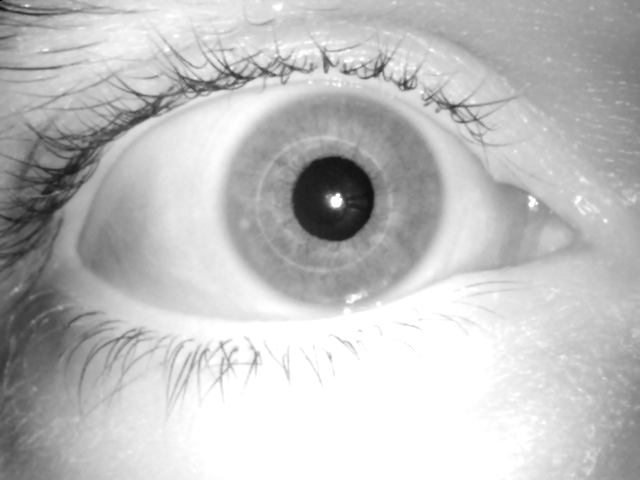}
\includegraphics[scale=0.16]{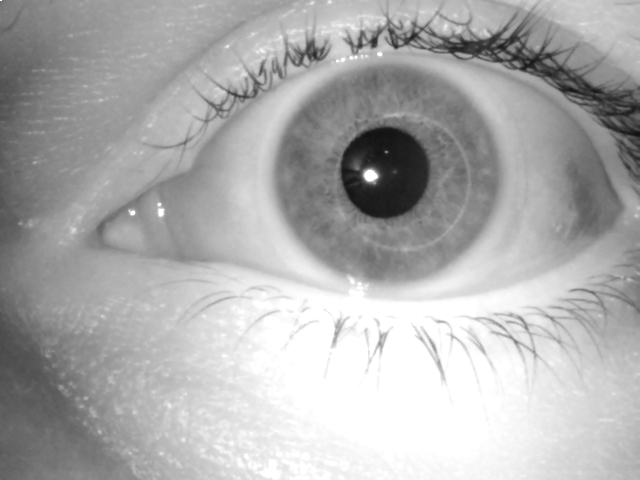}
\\
\includegraphics[scale=0.16]{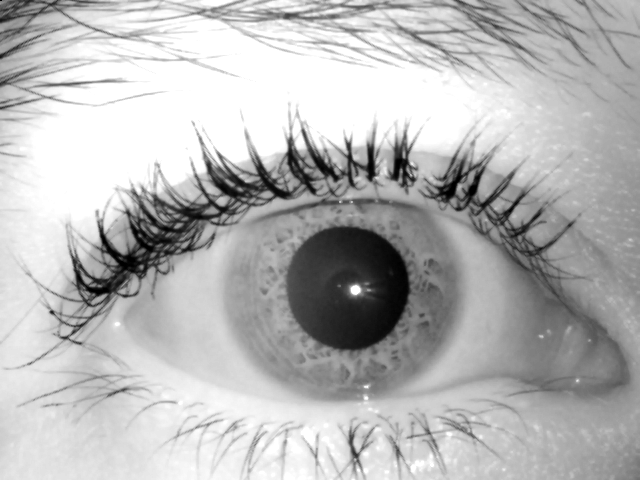}
\includegraphics[scale=0.16]{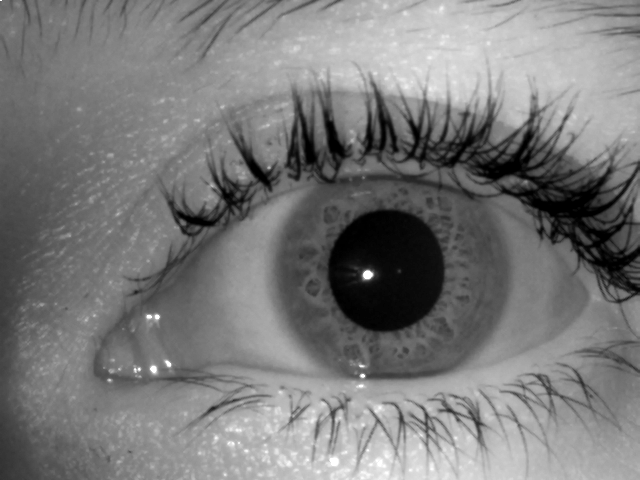}
\\
\includegraphics[scale=0.16]{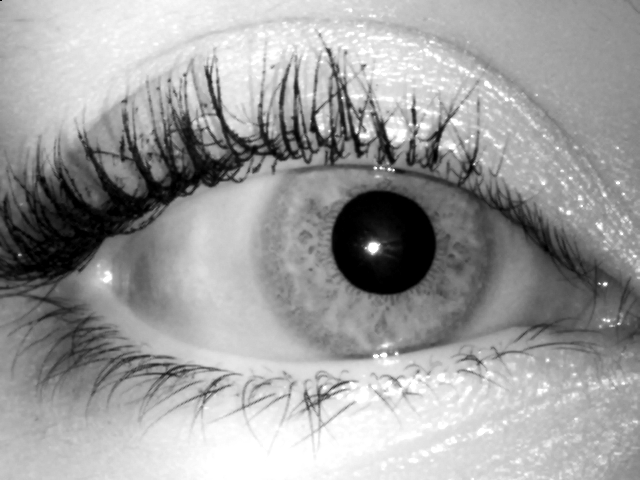}
\includegraphics[scale=0.16]{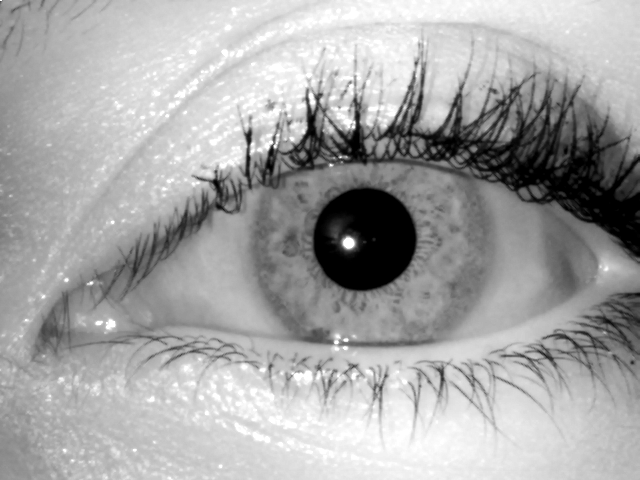}
\\
\includegraphics[scale=0.16]{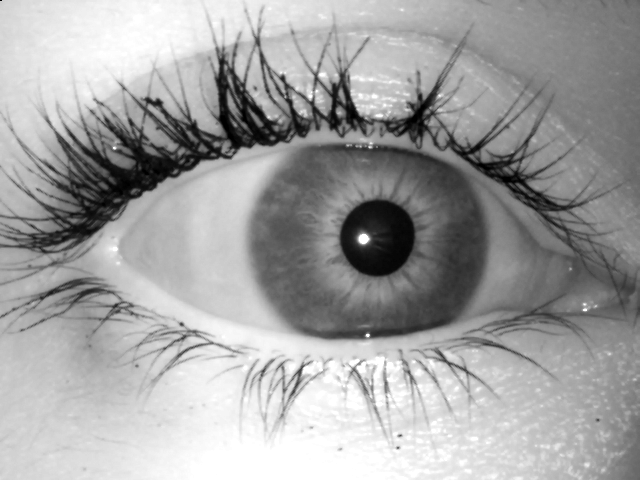}
\includegraphics[scale=0.16]{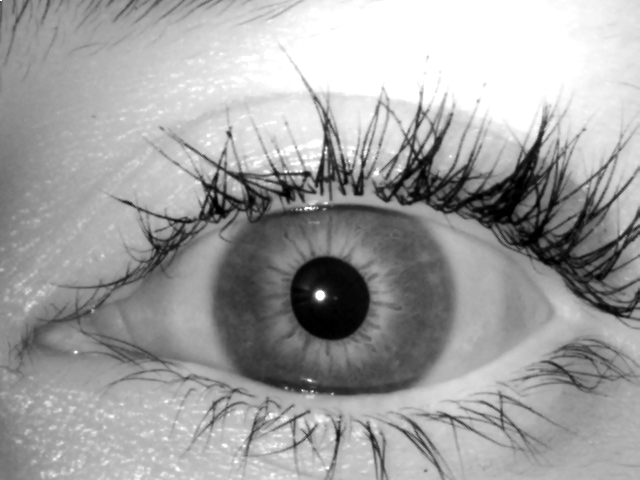}
\caption{Examples of the NIR images captured from top to bottom. a) Control, b) Alcohol, c) drug and d) sleepiness images.}
\label{img_examples}
\end{figure}

During the image acquisition process, subjects were positioned in front of the capture device, which detected the eyes and initiated the recording. Examples of images captured by the iris sensor are depicted in Figure~\ref{img_examples}.

The database has four classes of NIR image sequences in different conditions as follows:
~\textit{Control}: subjects that are not under alcohol and/or drug influence and in normal sleeping conditions (self-declared).
~\textit{Alcohol}: subjects who have consumed alcohol previously.
~\textit{Drugs}: subjects who consumed some drugs (mainly marijuana) or psychotropic drugs (by medical prescription).
~\textit{Sleep}: subjects with sleep deprivation due to sleep disorders related to occupational factors (shift structures).

In this work, the four classes were represented as two classes: FIT (control subject) and Unfit (Alcohol, drugs and sleepiness), which were grouped together.

\subsection{Metrics}
\label{sec:metric}

The False Positive Rate~(FPR) and False Negative Rate~(FNR) are categorised as Type I and Type II errors, respectively. These metrics effectively measure the extent to which the algorithm misclassifies images as either fit or unfit in terms of alcohol, drugs, and sleepiness. Both the FPR and FNR are influenced by the decision threshold.

A Detection Error Trade-off~(DET) curve is also provided for all the experiments. The Equal Error Rate~(EER) on the DET curve indicates the balance between the FPR and FNR; the values on this curve are expressed as percentages. Furthermore, two different operational points are identified: FNR\textsubscript{10}, which corresponds to an FPR fixed at 10\%, and FNR\textsubscript{20}, which occurs when the FPR is maintained at 5\%. Both FNR\textsubscript{10} and FNR\textsubscript{20} are independent of decision thresholds.

In terms of selecting the best image during the capture process, the ISO/IEC 29794-6\footnote{\url{https://www.iso.org/standard/54066.html}} standard on iris image quality defines a set of quality measures that can assess the utility of a sample. Based on the NIST-IREX evaluation\footnote{\url{https://www.nist.gov/programs-projects/iris-exchange-irex-overview}}, we employ a sharpness measure. Specifically, this work uses the Laplacian of Gaussian operator (LoG) as the sharpness measure. The sharpness of an image is determined by the intensity derived from filtering the image using a LoG kernel.

\begin{equation}
    LoG(x,y)=-\frac{1}{\pi\sigma^4} \left[1- \frac{x^2+y^2}{2\sigma^2}\right]e^{-\frac{x^2+y^2}{2\sigma^2}} \label{eq:sharp}
\end{equation}

The LoG is used in this work. The sharpness of an image is determined by the power of filtering the image. The LoG of~$(x,y)$ of an image with pixel intensity values $(x,y$)~is given by the standard deviation~$(\sigma)$~of the Gaussian is $1.4$. 

\section{Experiment and Results}
\label{sec:experiment}

\subsubsection{Preprocessing}
The original images have a size of $640\times480$. As a first step, all the single-eye images were resized to a $224\times224$ resolution based on the network CNN models. Both left and right eye images are processed in the same manner. Data augmentation techniques were applied based on the change of illumination, mirroring, blurring, 15º rotation and some artificial noise.

In order to study the influence of the foundation models, four experiments were defined based on training algorithms from scratch, zero-shot, applying fine-tuning in binary classification problems and the binary classification in a leave-one-out protocol. This protocol was applied to all the models, based on the CNN, VT, as well as two downstream models such as DinoV2 and OpenClip. For OpenClip, the pre-trained dataset "laion400m\_e32" was used.

\subsection{Experiment 1 - from scratch}

This experiment creates a baseline of different network implementations of CNN training from scratch using NIR images. The CNN models were MobileNetV3\_large~\cite{mbv3}, ResNet34~\cite{resnet}, ResNet101~\cite{resnet}, EfficientNetB0\-V2 \cite{efficientnet}, DenseNet121~\cite{DenseNet} and SwinTranformerV2~\cite{swin}. 
In case of SwinT models, the most detailed configuration is described. The patch size division that makes a stronger classifier with the following configuration: image-size = $224$, patch-size = $4$, num-channels = $3$, embed-dim = $96$, depths = [2, 2, 6, 2], num-heads = [3, 6, 12, 24], window-size = $7$, mlp-ratio = $4.0$, drop-path-rate = $0.1$, hidden-act = 'gelu', initializer-range = $0.02$, layer-norm-eps = $1e-05$, enc-stride = $3$. 

The parameters for all SwinT are selected according to:
SwinT-B, SwinT-S, and SwinT-L model sizes, whose complexities are increased from baseline SwinT-B by $\times0.5$,$\times1$, and $\times2$. Window size $M$ is set to $7$ by default, and the query dimension of each head(as multi-head attention is used) is $32$ for all these models.

\begin{figure*}[]
      \centering
      \includegraphics[scale=0.25]{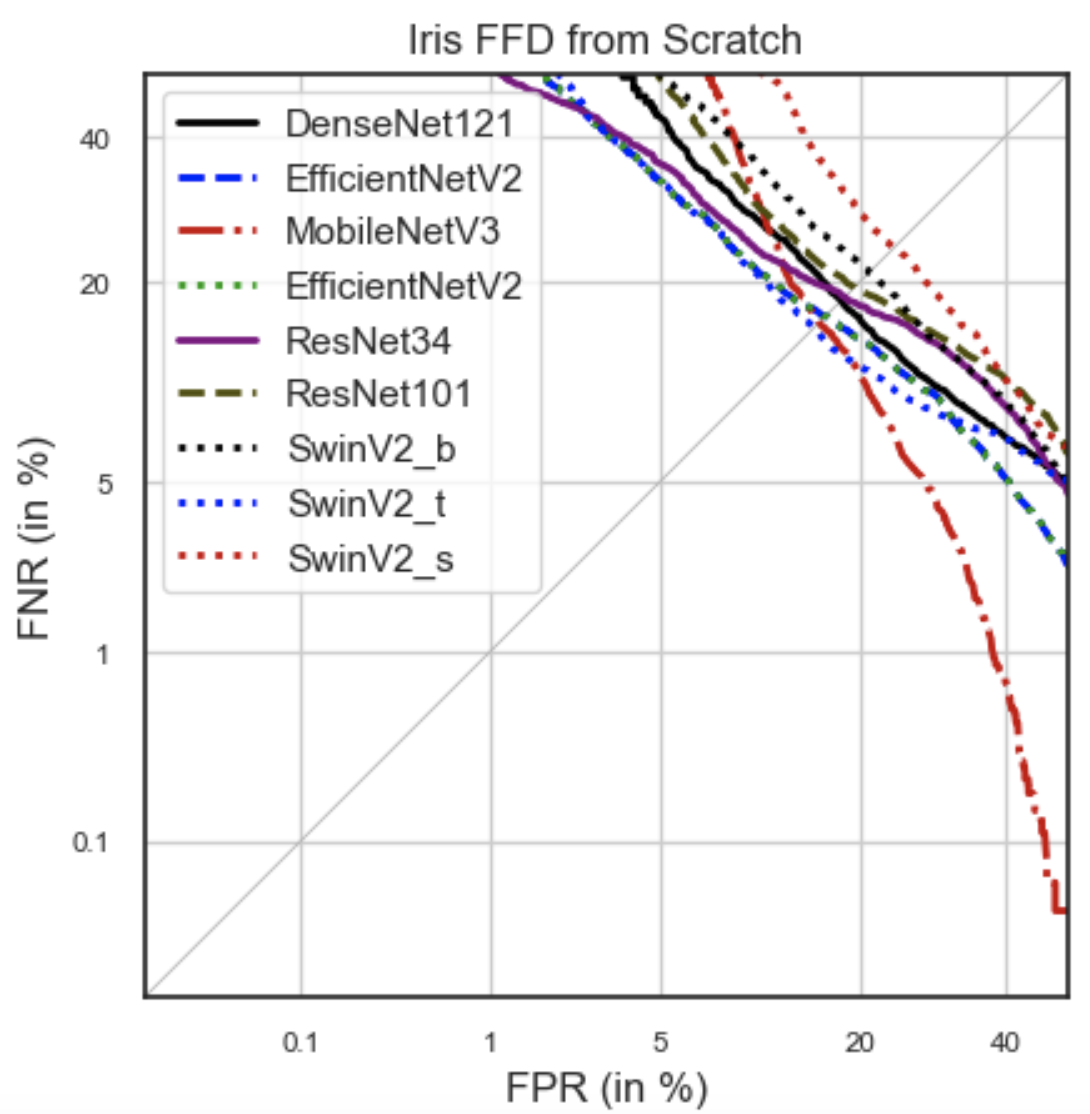}
      \includegraphics[scale=0.25]{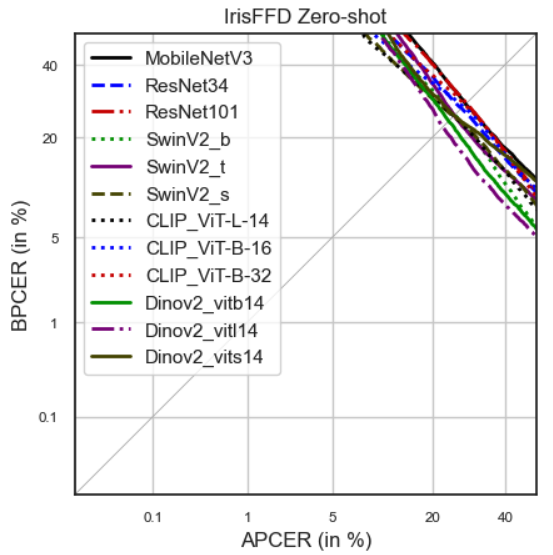}
      \includegraphics[scale=0.25]{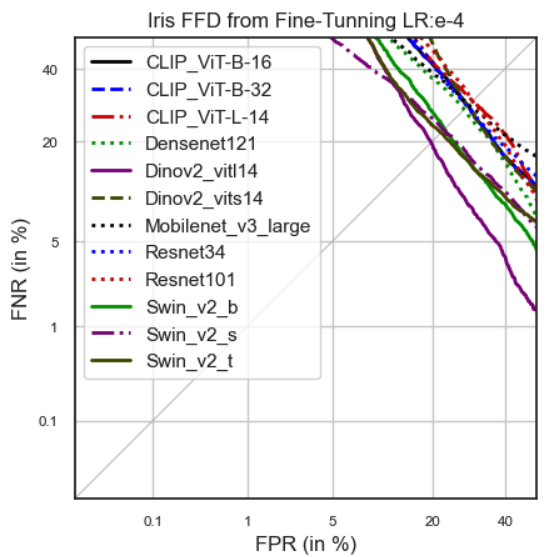}      
      \caption{\label{fig:alldet} DET curve results for Experiments 1 up to 3 for a binary classification problem (Fit/Unfit) using DL, VT and FM. Left to right: From Scratch, Zero-Shot and Fine-tuning.}
\end{figure*}

Figure \ref{fig:alldet} depicts the DET curves for Experiment 1 and other experiments in order to facilitate direct comparison and visualisation. Figure \ref{fig:alldet} also displays that for experiment 1, Swinv2\_t reached the best results with an EER of 15.34\%. 


\begin{table*}[]
\scriptsize
\centering
\caption{Summary results for Experiments 1 up to 3. N/A: Not applicable to FM.}
\label{tab:fit-all-Exp1-3}
\resizebox{\textwidth}{!}{%
\begin{tabular}{ccccccccccccc}
\hline
\multicolumn{1}{c|}{\multirow{2}{*}{Models}} &
  \multicolumn{4}{c|}{Scratch} &
  \multicolumn{4}{c|}{Zero-Shot} &
  \multicolumn{4}{c}{Fine-tunning} \\ \cline{2-13} 
\multicolumn{1}{c|}{} &
  \multicolumn{1}{c|}{\begin{tabular}[c]{@{}c@{}}EER\\ (\%)\end{tabular}} &
  \multicolumn{1}{c|}{\begin{tabular}[c]{@{}c@{}}FNMR\\ @10\\ (\%)\end{tabular}} &
  \multicolumn{1}{c|}{\begin{tabular}[c]{@{}c@{}}FNMR\\ @20\\ (\%)\end{tabular}} &
  \multicolumn{1}{c|}{\begin{tabular}[c]{@{}c@{}}FNMR\\ @100\\ (\%)\end{tabular}} &
  \multicolumn{1}{c|}{\begin{tabular}[c]{@{}c@{}}EER\\ (\%)\end{tabular}} &
  \multicolumn{1}{c|}{\begin{tabular}[c]{@{}c@{}}FNMR\\ @10\\ (\%)\end{tabular}} &
  \multicolumn{1}{c|}{\begin{tabular}[c]{@{}c@{}}FNMR\\ @20\\ (\%)\end{tabular}} &
  \multicolumn{1}{c|}{\begin{tabular}[c]{@{}c@{}}FNMR\\ @100\\ (\%)\end{tabular}} &
  \multicolumn{1}{l|}{\begin{tabular}[c]{@{}l@{}}EER\\ (\%)\end{tabular}} &
  \multicolumn{1}{c|}{\begin{tabular}[c]{@{}c@{}}FNMR\\ @10\\ (\%)\end{tabular}} &
  \multicolumn{1}{c|}{\begin{tabular}[c]{@{}c@{}}FNMR\\ @20\\ (\%)\end{tabular}} &
  \multicolumn{1}{c|}{\begin{tabular}[c]{@{}c@{}}FNMR\\ @100\\ (\%)\end{tabular}} \\ \hline
\multicolumn{13}{c}{Foundation Models} \\ \hline
\multicolumn{1}{c|}{CLIP\_ViT-B16} &
  \multicolumn{4}{c|}{\multirow{6}{*}{N/A}} &
  \multicolumn{1}{c|}{27.71} &
  \multicolumn{1}{c|}{48.51} &
  \multicolumn{1}{c|}{61.38} &
  \multicolumn{1}{c|}{85.38} &
  \multicolumn{1}{c|}{30.09} &
  \multicolumn{1}{c|}{66.46} &
  \multicolumn{1}{c|}{81.29} &
  94.24 \\ \cline{1-1} \cline{6-13} 
\multicolumn{1}{c|}{CLIP\_ViT-B32} &
  \multicolumn{4}{c|}{} &
  \multicolumn{1}{c|}{28.19} &
  \multicolumn{1}{c|}{53.27} &
  \multicolumn{1}{c|}{68.16} &
  \multicolumn{1}{c|}{88.33} &
  \multicolumn{1}{c|}{28.97} &
  \multicolumn{1}{c|}{59.09} &
  \multicolumn{1}{c|}{72.74} &
  90.52 \\ \cline{1-1} \cline{6-13} 
\multicolumn{1}{c|}{CLIP\_ViT-L14} &
  \multicolumn{4}{c|}{} &
  \multicolumn{1}{c|}{25.18} &
  \multicolumn{1}{c|}{44.75} &
  \multicolumn{1}{c|}{59.15} &
  \multicolumn{1}{c|}{82.39} &
  \multicolumn{1}{c|}{30.25} &
  \multicolumn{1}{c|}{67.93} &
  \multicolumn{1}{c|}{80.43} &
  95.01 \\ \cline{1-1} \cline{6-13} 
\multicolumn{1}{c|}{DinoV2-B} &
  \multicolumn{4}{c|}{} &
  \multicolumn{1}{c|}{23.91} &
  \multicolumn{1}{c|}{51.48} &
  \multicolumn{1}{c|}{69.98} &
  \multicolumn{1}{c|}{99.93} &
  \multicolumn{1}{c|}{23.61} &
  \multicolumn{1}{c|}{46.59} &
  \multicolumn{1}{c|}{65.20} &
  83.70 \\ \cline{1-1} \cline{6-13} 
\multicolumn{1}{c|}{DinoV2-S} &
  \multicolumn{4}{c|}{} &
  \multicolumn{1}{c|}{25.33} &
  \multicolumn{1}{c|}{56.21} &
  \multicolumn{1}{c|}{75.18} &
  \multicolumn{1}{c|}{95.74} &
  \multicolumn{1}{c|}{29.91} &
  \multicolumn{1}{c|}{71.82} &
  \multicolumn{1}{c|}{82.53} &
  94.73 \\ \cline{1-1} \cline{6-13} 
\multicolumn{1}{c|}{DinoV2-L} &
  \multicolumn{4}{c|}{} &
  \multicolumn{1}{c|}{\mycc22.37} &
  \multicolumn{1}{c|}{\mycc51.13} &
  \multicolumn{1}{c|}{\mycc72.87} &
  \multicolumn{1}{c|}{\mycc93.45} &
  \multicolumn{1}{c|}{\mycc19.76} &
  \multicolumn{1}{c|}{\mycc43.74} &
  \multicolumn{1}{c|}{\mycc66.18} &
  \mycc88.19 \\ \hline
\multicolumn{13}{c}{Deep Learning Models} \\ \hline
\multicolumn{1}{c|}{DenseNet} &
  \multicolumn{1}{c|}{17.97} &
  \multicolumn{1}{c|}{28.55} &
  \multicolumn{1}{c|}{43.01} &
  \multicolumn{1}{c|}{53.06} &
  \multicolumn{1}{c|}{27.69} &
  \multicolumn{1}{c|}{56.99} &
  \multicolumn{1}{c|}{68.36} &
  \multicolumn{1}{c|}{88.09} &
  \multicolumn{1}{c|}{27.88} &
  \multicolumn{1}{c|}{53.64} &
  \multicolumn{1}{c|}{67.53} &
  85.65 \\ \hline
\multicolumn{1}{c|}{MobileNetV3} &
  \multicolumn{1}{c|}{15.78} &
  \multicolumn{1}{c|}{32.81} &
  \multicolumn{1}{c|}{56.18} &
  \multicolumn{1}{c|}{56.18} &
  \multicolumn{1}{c|}{28.36} &
  \multicolumn{1}{c|}{60.71} &
  \multicolumn{1}{c|}{74.25} &
  \multicolumn{1}{c|}{91.33} &
  \multicolumn{1}{c|}{29.13} &
  \multicolumn{1}{c|}{52.83} &
  \multicolumn{1}{c|}{61.61} &
  79.22 \\ \hline
\multicolumn{1}{c|}{ResNet34} &
  \multicolumn{1}{c|}{18.38} &
  \multicolumn{1}{c|}{24.64} &
  \multicolumn{1}{c|}{35.77} &
  \multicolumn{1}{c|}{52.33} &
  \multicolumn{1}{c|}{27.25} &
  \multicolumn{1}{c|}{53.57} &
  \multicolumn{1}{c|}{67.57} &
  \multicolumn{1}{c|}{86.22} &
  \multicolumn{1}{c|}{28.84} &
  \multicolumn{1}{c|}{59.28} &
  \multicolumn{1}{c|}{71.74} &
  84.47 \\ \hline
\multicolumn{1}{c|}{ResNet101} &
  \multicolumn{1}{c|}{19.43} &
  \multicolumn{1}{c|}{30.88} &
  \multicolumn{1}{c|}{49.19} &
  \multicolumn{1}{c|}{64.28} &
  \multicolumn{1}{c|}{28.54} &
  \multicolumn{1}{c|}{58.54} &
  \multicolumn{1}{c|}{73.68} &
  \multicolumn{1}{c|}{91.41} &
  \multicolumn{1}{c|}{30.28} &
  \multicolumn{1}{c|}{64.26} &
  \multicolumn{1}{c|}{76.99} &
  91.03 \\ \hline
\multicolumn{13}{|c|}{Vision Transformers} \\ \hline
\multicolumn{1}{c|}{SwinT-V2-B} &
  \multicolumn{1}{c|}{21.09} &
  \multicolumn{1}{c|}{37.29} &
  \multicolumn{1}{c|}{50.03} &
  \multicolumn{1}{c|}{69.11} &
  \multicolumn{1}{c|}{25.50} &
  \multicolumn{1}{c|}{51.30} &
  \multicolumn{1}{c|}{67.59} &
  \multicolumn{1}{c|}{86.77} &
  \multicolumn{1}{c|}{21.43} &
  \multicolumn{1}{c|}{41.24} &
  \multicolumn{1}{c|}{56.40} &
  73.90 \\ \hline
\multicolumn{1}{c|}{SwinT-V-T} &
  \multicolumn{1}{c|}{ \mycc15.34} &
  \multicolumn{1}{c|}{\mycc21.74} &
  \multicolumn{1}{c|}{\mycc33.08} &
  \multicolumn{1}{c|}{\mycc65.20} &
  \multicolumn{1}{c|}{25.32} &
  \multicolumn{1}{c|}{59.28} &
  \multicolumn{1}{c|}{75.62} &
  \multicolumn{1}{c|}{94.83} &
  \multicolumn{1}{c|}{21.55} &
  \multicolumn{1}{c|}{44.19} &
  \multicolumn{1}{c|}{68.81} &
  88.64 \\ \hline
\multicolumn{1}{c|}{SwinTV2-S} &
  \multicolumn{1}{c|}{24.10} &
  \multicolumn{1}{c|}{51.67} &
  \multicolumn{1}{c|}{67.46} &
  \multicolumn{1}{c|}{80.78} &
  \multicolumn{1}{c|}{25.03} &
  \multicolumn{1}{c|}{46.05} &
  \multicolumn{1}{c|}{61.53} &
  \multicolumn{1}{c|}{88.83} &
  \multicolumn{1}{c|}{22.68} &
  \multicolumn{1}{c|}{38.32} &
  \multicolumn{1}{c|}{48.39} &
  66.33 \\ \hline
\end{tabular}%
}
\end{table*}

The results show that the FFD is a challenging problem based on traditional CNNs and the powerful vision Transformer method used to classify this control group. We believe that the lack of images makes a difference in this approach.

For these tasks all the models were trained for 50 epochs and a learning rate of $1e-4$, using an Adam optimizer.

\subsection{Experiment 2 - Zero-Shot}

In order to compare the results with experiment 1, we analysed the potential of foundation models for FFD. This experiment explored several variations of the DinoV2 self-supervised vision transformers' backbones, such as DinoV2-ViTS14, DinoV2-ViTB14 and DinoV2-ViTL. Each one of these models contains 22.2, 86.9 and 304.0 million parameters, respectively. 

Further on, we also explored OpenClip models, such as VIT-B-16 with 86.3, ViT-B-32 with 88.0 and ViT-L-14 with 304 million parameters. In all cases, we only used the visual encoder and removed the text encoder.

In the Zero-shot approach, we assess all the models from Deep learning (DL), Vision transformer (VT) and Foundation models (FM) under the condition that all the layers were frozen. Thus, each network was used as a feature extractor with different embedding sizes that went through a sigmoid for the DL and the VT and a Gelu for the Dinov2 and OpenClip. The DinoV2 and OpenClip models were used to extract the features of the size $1\times768$ per image.

Figure \ref{fig:alldet} depicts the DET curves for Experiment 2 and other experiments in order to facilitate direct comparison and visualisation. Figure \ref{fig:alldet} also displays that Dinov2-ViT-L-14 reached the best results with an EER of 22.37\%. 

\begin{table*}[]
\caption{Summary results LOO protocol. ACC represents the accuracy.}
\label{tab:fit-all-loo}
\resizebox{\textwidth}{!}{%
\begin{tabular}{ccccccccccclllll}
\hline
\multicolumn{1}{c|}{\multirow{2}{*}{Models}} &
  \multicolumn{5}{c|}{\begin{tabular}[c]{@{}c@{}}LOO\\ Train: Control vs Drug-Sleepiness\\ Test: Alcohol\end{tabular}} &
  \multicolumn{5}{c|}{\begin{tabular}[c]{@{}c@{}}LOO\\ Train: Control vs Alcohol-Sleepiness\\ Test: Drug\end{tabular}} &
  \multicolumn{5}{c}{\begin{tabular}[c]{@{}c@{}}LOO\\ Train: Control vs Drug-Alcohol\\ Test Sleepiness\end{tabular}} \\ \cline{2-16} 
\multicolumn{1}{c|}{} &
  \multicolumn{1}{c|}{\begin{tabular}[c]{@{}c@{}}EER\\ (\%)\end{tabular}} &
  \multicolumn{1}{c|}{\begin{tabular}[c]{@{}c@{}}FNMR\\ @10\\ (\%)\end{tabular}} &
  \multicolumn{1}{c|}{\begin{tabular}[c]{@{}c@{}}FNMR\\ @20\\ (\%)\end{tabular}} &
  \multicolumn{1}{c|}{\begin{tabular}[c]{@{}c@{}}FNMR\\ @100\\ (\%)\end{tabular}} &
  \multicolumn{1}{c|}{\begin{tabular}[c]{@{}c@{}}ACC\\ (\%)\end{tabular}} &
  \multicolumn{1}{c|}{\begin{tabular}[c]{@{}c@{}}EER\\ (\%)\end{tabular}} &
  \multicolumn{1}{c|}{\begin{tabular}[c]{@{}c@{}}FNMR\\ @10\\ (\%)\end{tabular}} &
  \multicolumn{1}{c|}{\begin{tabular}[c]{@{}c@{}}FNMR\\ @20\\ (\%)\end{tabular}} &
  \multicolumn{1}{c|}{\begin{tabular}[c]{@{}c@{}}FNMR\\ @100\\ (\%)\end{tabular}} &
  \multicolumn{1}{c|}{\begin{tabular}[c]{@{}c@{}}ACC\\ (\%)\end{tabular}} &
  \multicolumn{1}{l|}{\begin{tabular}[c]{@{}l@{}}EER\\ (\%)\end{tabular}} &
  \multicolumn{1}{c|}{\begin{tabular}[c]{@{}c@{}}FNMR\\ @10\\ (\%)\end{tabular}} &
  \multicolumn{1}{c|}{\begin{tabular}[c]{@{}c@{}}FNMR\\ @20\\ (\%)\end{tabular}} &
  \multicolumn{1}{c|}{\begin{tabular}[c]{@{}c@{}}FNMR\\ @100\\ (\%)\end{tabular}} &
  \multicolumn{1}{c}{\begin{tabular}[c]{@{}c@{}}ACC\\ (\%)\end{tabular}} \\ \hline
\multicolumn{16}{c}{Foundation Models} \\ \hline
\multicolumn{1}{c|}{DinoV2-B} &
  \multicolumn{1}{c|}{3.16} &
  \multicolumn{1}{c|}{0.92} &
  \multicolumn{1}{c|}{2.64} &
  \multicolumn{1}{c|}{6.34} &
  \multicolumn{1}{c|}{96.09} &
  \multicolumn{1}{c|}{\mycc1.80} &
  \multicolumn{1}{c|}{\mycc0.52} &
  \multicolumn{1}{c|}{\mycc1.45} &
  \multicolumn{1}{c|}{\mycc2.37} &
  \multicolumn{1}{c|}{\mycc97.34} &
  \multicolumn{1}{l|}{\mycc2.13} &
  \multicolumn{1}{l|}{\mycc0.52} &
  \multicolumn{1}{l|}{\mycc1.18} &
  \multicolumn{1}{l|}{\mycc14.01} &
  \mycc98.83 \\ \hline
\multicolumn{1}{c|}{DinoV2-S} &
  \multicolumn{1}{c|}{\mycc2.89} &
  \multicolumn{1}{c|}{\mycc0.79} &
  \multicolumn{1}{c|}{\mycc2.11} &
  \multicolumn{1}{c|}{\mycc20.67} &
  \multicolumn{1}{c|}{\mycc95.48} &
  \multicolumn{1}{c|}{5.07} &
  \multicolumn{1}{c|}{2.11} &
  \multicolumn{1}{c|}{5.68} &
  \multicolumn{1}{c|}{39.23} &
  \multicolumn{1}{c|}{96.43} &
  \multicolumn{1}{l|}{3.44} &
  \multicolumn{1}{l|}{0.79} &
  \multicolumn{1}{l|}{2.50} &
  \multicolumn{1}{l|}{21.53} &
  98.72 \\ \hline
\multicolumn{1}{c|}{DinoV2-L} &
  \multicolumn{1}{c|}{3.55} &
  \multicolumn{1}{c|}{1.84} &
  \multicolumn{1}{c|}{3.17} &
  \multicolumn{1}{c|}{18.94} &
  \multicolumn{1}{c|}{95.48} &
  \multicolumn{1}{c|}{3.03} &
  \multicolumn{1}{c|}{1.71} &
  \multicolumn{1}{c|}{2.11} &
  \multicolumn{1}{c|}{4.62} &
  \multicolumn{1}{c|}{95.76} &
  \multicolumn{1}{l|}{10.23} &
  \multicolumn{1}{l|}{15.05} &
  \multicolumn{1}{l|}{15.85} &
  \multicolumn{1}{l|}{16.90} &
  98.24 \\ \hline
\multicolumn{16}{c}{Deep Learning Models} \\ \hline
\multicolumn{1}{c|}{DenseNet} &
  \multicolumn{1}{c|}{11.39} &
  \multicolumn{1}{c|}{18.49} &
  \multicolumn{1}{c|}{30.51} &
  \multicolumn{1}{c|}{51.38} &
  \multicolumn{1}{c|}{84.82} &
  \multicolumn{1}{c|}{6.74} &
  \multicolumn{1}{c|}{4.09} &
  \multicolumn{1}{c|}{15.19} &
  \multicolumn{1}{c|}{37.51} &
  \multicolumn{1}{c|}{87.18} &
  \multicolumn{1}{l|}{6.20} &
  \multicolumn{1}{c|}{3.42} &
  \multicolumn{1}{c|}{12.41} &
  \multicolumn{1}{c|}{31.71} &
  \multicolumn{1}{c}{91.66} \\ \hline
\multicolumn{1}{c|}{EfficientNetV2} &
  \multicolumn{1}{c|}{11.89} &
  \multicolumn{1}{c|}{13.87} &
  \multicolumn{1}{c|}{19.15} &
  \multicolumn{1}{c|}{45.04} &
  \multicolumn{1}{c|}{80.50} &
  \multicolumn{1}{c|}{8.37} &
  \multicolumn{1}{c|}{7.52} &
  \multicolumn{1}{c|}{17.17} &
  \multicolumn{1}{c|}{26.85} &
  \multicolumn{1}{c|}{82.63} &
  \multicolumn{1}{l|}{6.75} &
  \multicolumn{1}{l|}{5.41} &
  \multicolumn{1}{l|}{8.19} &
  \multicolumn{1}{l|}{27.08} &
  87.07 \\ \hline
\multicolumn{1}{c|}{MobileNetV3} &
  \multicolumn{1}{c|}{13.87} &
  \multicolumn{1}{c|}{22.06} &
  \multicolumn{1}{c|}{25.75} &
  \multicolumn{1}{c|}{39.49} &
  \multicolumn{1}{c|}{85.22} &
  \multicolumn{1}{c|}{10.03} &
  \multicolumn{1}{c|}{10.03} &
  \multicolumn{1}{c|}{20.73} &
  \multicolumn{1}{c|}{31.96} &
  \multicolumn{1}{c|}{86.01} &
  \multicolumn{1}{l|}{11.67} &
  \multicolumn{1}{l|}{17.43} &
  \multicolumn{1}{l|}{19.15} &
  \multicolumn{1}{l|}{21.79} &
  91.32 \\ \hline
\multicolumn{1}{c|}{ResNet34} &
  \multicolumn{1}{c|}{8.61} &
  \multicolumn{1}{c|}{8.32} &
  \multicolumn{1}{c|}{11.49} &
  \multicolumn{1}{c|}{20.34} &
  \multicolumn{1}{c|}{83.63} &
  \multicolumn{1}{c|}{8.54} &
  \multicolumn{1}{c|}{8.32} &
  \multicolumn{1}{c|}{10.43} &
  \multicolumn{1}{c|}{14.39} &
  \multicolumn{1}{c|}{86.40} &
  \multicolumn{1}{l|}{11.26} &
  \multicolumn{1}{l|}{15.58} &
  \multicolumn{1}{l|}{18.89} &
  \multicolumn{1}{l|}{24.24} &
  91.09 \\ \hline
\multicolumn{1}{c|}{ResNet101} &
  \multicolumn{1}{c|}{11.80} &
  \multicolumn{1}{c|}{28.53} &
  \multicolumn{1}{c|}{36.85} &
  \multicolumn{1}{c|}{56.53} &
  \multicolumn{1}{c|}{84.70} &
  \multicolumn{1}{c|}{11.80} &
  \multicolumn{1}{c|}{28.53} &
  \multicolumn{1}{c|}{36.85} &
  \multicolumn{1}{c|}{56.53} &
  \multicolumn{1}{c|}{84.70} &
  \multicolumn{1}{l|}{10.32} &
  \multicolumn{1}{l|}{15.85} &
  \multicolumn{1}{l|}{26.15} &
  \multicolumn{1}{l|}{54.29} &
  91.28 \\ \hline
\multicolumn{16}{c}{Vision Transformers} \\ \hline
\multicolumn{1}{c|}{SwinT-V2-B} &
  \multicolumn{1}{c|}{17.01} &
  \multicolumn{1}{c|}{18.89} &
  \multicolumn{1}{c|}{21.79} &
  \multicolumn{1}{c|}{30.51} &
  \multicolumn{1}{c|}{81.91} &
  \multicolumn{1}{c|}{12.92} &
  \multicolumn{1}{c|}{16.11} &
  \multicolumn{1}{c|}{16.90} &
  \multicolumn{1}{c|}{23.77} &
  \multicolumn{1}{c|}{85.34} &
  \multicolumn{1}{l|}{16.03} &
  \multicolumn{1}{l|}{16.38} &
  \multicolumn{1}{l|}{17.04} &
  \multicolumn{1}{l|}{18.89} &
  89.91 \\ \hline
\multicolumn{1}{c|}{SwinT-V-L} &
  \multicolumn{1}{c|}{17.56} &
  \multicolumn{1}{c|}{25.49} &
  \multicolumn{1}{c|}{42.14} &
  \multicolumn{1}{c|}{64.72} &
  \multicolumn{1}{c|}{81.55} &
  \multicolumn{1}{c|}{16.10} &
  \multicolumn{1}{c|}{17.04} &
  \multicolumn{1}{c|}{22.60} &
  \multicolumn{1}{c|}{46.89} &
  \multicolumn{1}{c|}{84.89} &
  \multicolumn{1}{l|}{16.23} &
  \multicolumn{1}{l|}{16.38} &
  \multicolumn{1}{l|}{17.43} &
  \multicolumn{1}{l|}{38.83} &
  89.41 \\ \hline
\multicolumn{1}{c|}{SwinTV2-S} &
  \multicolumn{1}{c|}{19.79} &
  \multicolumn{1}{c|}{24.04} &
  \multicolumn{1}{c|}{30.28} &
  \multicolumn{1}{c|}{59.84} &
  \multicolumn{1}{c|}{81.26} &
  \multicolumn{1}{c|}{10.03} &
  \multicolumn{1}{c|}{10.03} &
  \multicolumn{1}{c|}{21.13} &
  \multicolumn{1}{c|}{41.08} &
  \multicolumn{1}{c|}{84.78} &
  \multicolumn{1}{l|}{16.23} &
  \multicolumn{1}{l|}{17.17} &
  \multicolumn{1}{l|}{17.70} &
  \multicolumn{1}{l|}{20.60} &
  89.06 \\ \hline
\end{tabular}%
}
\end{table*}

\begin{figure*}[]
      \centering
      \includegraphics[scale=0.28]{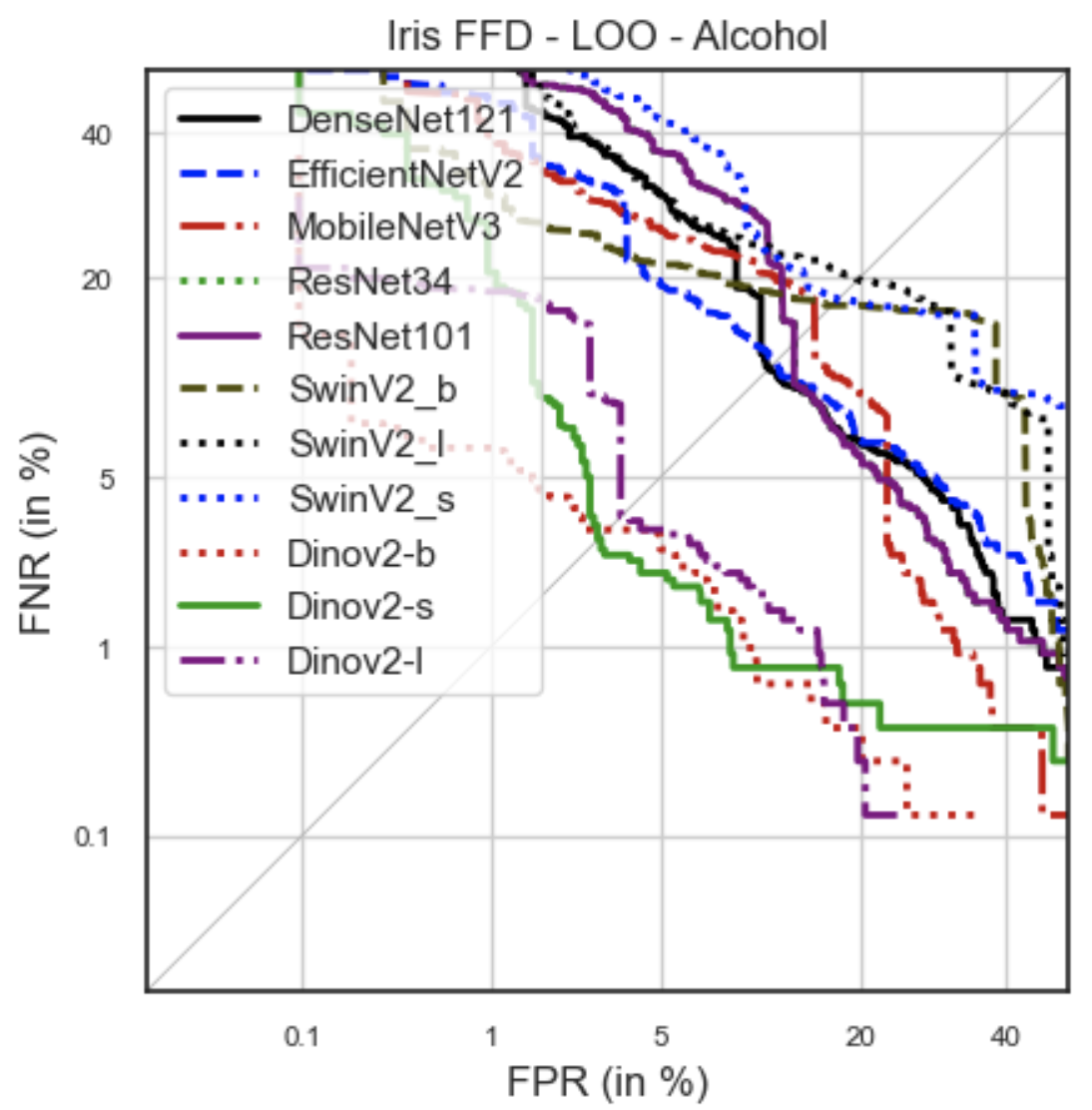}
      \includegraphics[scale=0.28]{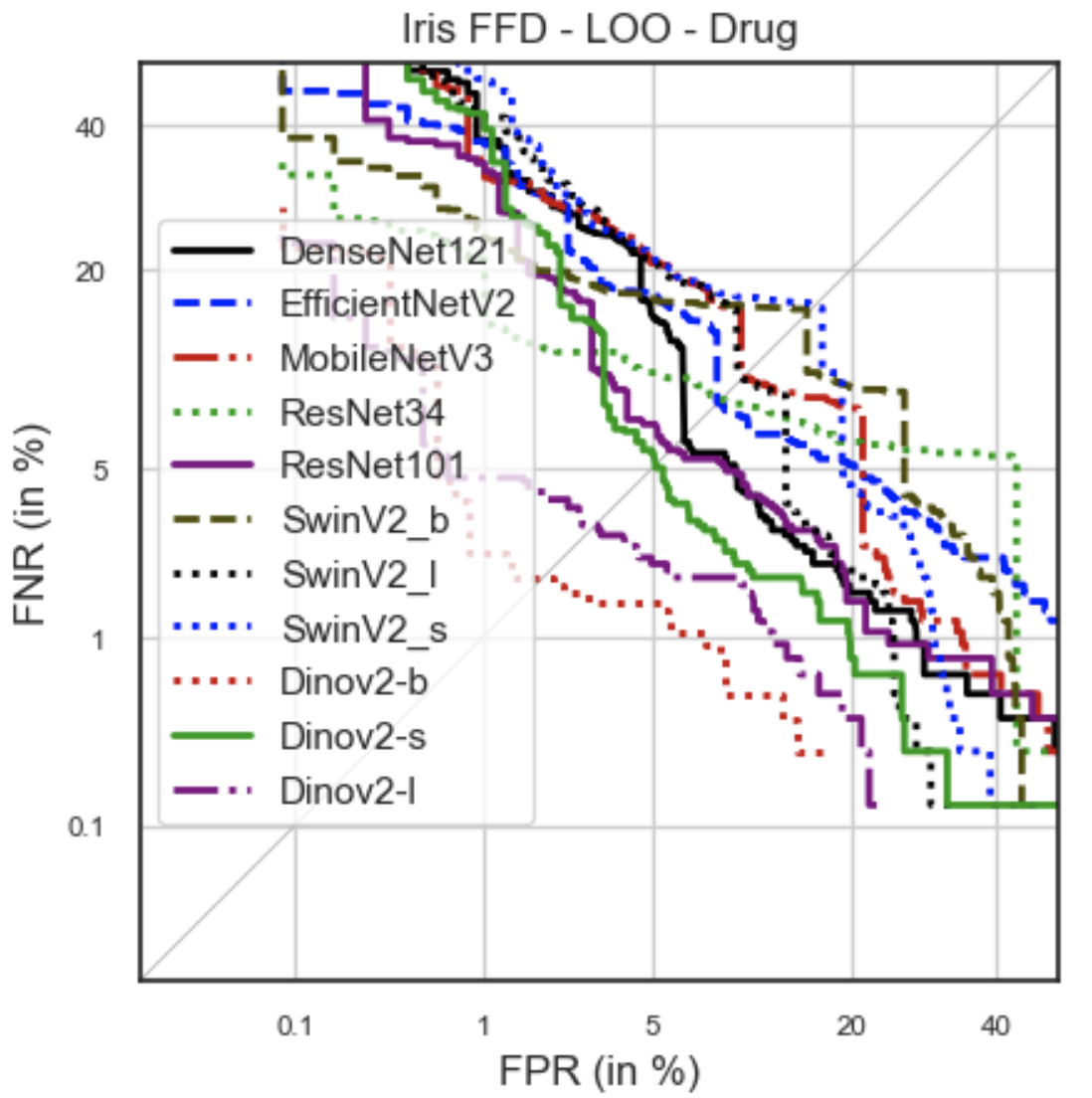}
      \includegraphics[scale=0.28]{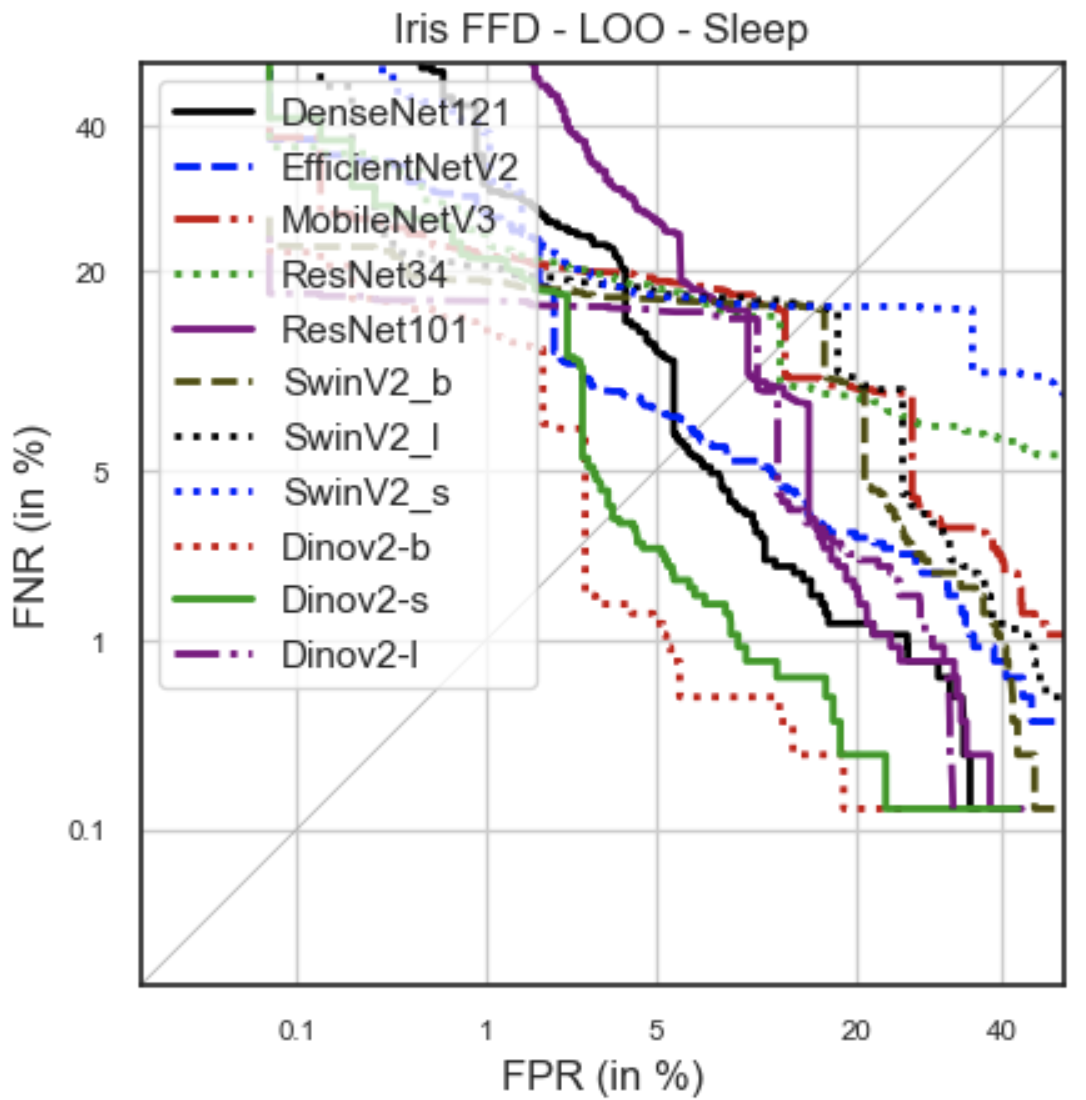}
      \caption{Experiment 3, DET curve for LOO protocol.  From left to right: Alcohol, Drug and Sleep using a learning rate of $1e-4$.}
      \label{fig:fit-loo}
\end{figure*}

\subsection{Experiment 3 - Fine-tuning}

The Dinov2 and OpenClip models were used to extract the features of the size $1\times768$ per image. Afterwards, this embedding is sent to a small head based on a neural network with three layers of size of $1 \times 364$,  $1 \times 182$ and $1 \times 64$ and Gaussian Error Units (GELU)~\cite{Gelu} activations between the layers. A grid search method was implemented in order to determine the best number of layers. The best results were obtained with one layer and GELU. The learning rate was explored in a grid search. The best value was achieved with $ 10^ {- 4} $. The CNN models were used with all layers frozen in order to obtain a representative embedding before entering the small neural network. These pre-trained models are based on ImageNet weight \cite{imagenet}. We applied the same conditions and parameters to DinoV2 and OpenClip in the fine-tuning process. All the models were trained for 50 epochs using an Adam optimiser on an A100 Nvidia GPU. 

Figure \ref{fig:alldet} shows the DET curve for all the traditional CNN models, together with all the variants of DinoV2 and OpenClip with different model sizes. The best result was reached by DinoV2 version ViT-L-14 on a learning rate of $1e-4$, reaching the best results with an EER of 19.76\%. Furthermore, we can observe that the foundation models reached the best results.

The Table~\ref{tab:fit-all-Exp1-3} shows the Equal Error Rate, FNR10, FNR20 and FNR100 for the experiment from 1 up to 3. 

\subsection{Experiment 4 - Leave-One-Out protocol}

This experiment analyses all the Leave-One-Out protocol~(LOO) models for the foundation model that obtained the best results, which was DinoV2. This means that the classifier is used to predict control group images (FIT) versus drug and sleep images for training in the first round, and alcohol is used for testing. In the second round, again, control group images (FIT) versus alcohol and drug images are used for training and sleepiness for testing, and so on, up to completing all the rotations. 

This protocol is a very challenging evaluation method for estimating performance in an unknown dataset. On this problem, this unknown condition could be a person under some new external factor such as alcohol, drugs, or sleepiness. 

The Table~\ref{tab:fit-all-loo} shows the Equal Error Rate, FNR10, FNR20 and FNR100. The EER for Alcohol was 2.89\%, For drugs, the EER was 1.80\% and for sleepiness, the EER was 2.13\%. As we noted, the drug prediction achieved a higher performance (lower EER). 

As expected, this protocol was challenging for the CNN model based on an unknown subset. Surprisingly, this approach is favourable to DinoV2, which obtained very competitive results and outperformed all the previous methods. Alcohol, drugs and sleep are hard to predict as a unique class (unfit), but the absence of one of them helps to improve the training process, showing excellent generalisation capabilities.  

In Figure \ref{fig:fit-loo}, we show the DET curves for a LOO protocol. In all the conditions, the model based on DinoV2-B achieved the best results.

\subsection{Benchmark with the state of the art}

In order to benchmark our results with the state of the art, we can use the accuracy metric (ACC) according to the results reported in previous approaches. Table \ref{tab:comparison} shows that the foundation models reached competitive results and outperformed some previous implementations. It is relevant to highlight that all the methods are based on feature extraction with a classifier in the head. However, the embedding result from the foundation model yielded more discriminative features, allowing us to improve the results.

\begin{table}[H]
\caption{Summary Benchmark with state-of-the-art.}
\label{tab:comparison}
\resizebox{\columnwidth}{!}{%
\begin{tabular}{cccccc}
\hline
Method &
  \multicolumn{1}{l}{Cond} &
  \begin{tabular}[c]{@{}c@{}}Sensitivity\\ (\%)\end{tabular} &
  \begin{tabular}[c]{@{}c@{}}Specificity\\ (\%)\end{tabular} &
  \begin{tabular}[c]{@{}c@{}}F1-Score\\ (\%)\end{tabular} &
  \begin{tabular}[c]{@{}c@{}}Accuracy\\ (\%)\end{tabular} \\ \hline
\multirow{2}{*}{\begin{tabular}[c]{@{}c@{}}RF\\ {[}3{]}\end{tabular}}         & Fit   & 70.1 & 94.7 & 80.5 & \multirow{2}{*}{70.8}          \\ \cline{2-5}
                                                                              & Unfit & 75.2 & 28.5 & 41.3 &                                \\ \hline
\multirow{2}{*}{\begin{tabular}[c]{@{}c@{}}GMB\\ {[}3{]}\end{tabular}}        & Fit   & 73.1 & 95.8 & 82.9 & \multirow{2}{*}{73.1}          \\ \cline{2-5}
                                                                              & Unfit & 79.8 & 40.0 & 45.7 &                                \\ \hline
\multirow{2}{*}{\begin{tabular}[c]{@{}c@{}}MLP \\ {[}3{]}\end{tabular}}       & Fit   & 75.3 & 95.4 & 84.2 & \multirow{2}{*}{75.3}          \\ \cline{2-5}
                                                                              & Unfit & 77.1 & 33.1 & 46.3 &                                \\ \hline
\multirow{2}{*}{\begin{tabular}[c]{@{}c@{}}CNN-LSTM \\ {[}27{]}\end{tabular}} & Fit   & 81.4 & 99.3 & 89.5 & \multirow{2}{*}{83.6}          \\ \cline{2-5}
                                                                              & Unfit & 96.9 & 46.9 & 63.3 &                                \\ \hline
\multirow{2}{*}{\begin{tabular}[c]{@{}c@{}}CNN-Single-Frame\\ {[}24{]}\end{tabular}} &
  \multirow{2}{*}{\begin{tabular}[c]{@{}c@{}}Fit vs\\ Unfit\end{tabular}} &
  \multirow{2}{*}{98.75} &
  \multirow{2}{*}{66.0} &
  \multirow{2}{*}{90.5} &
  \multirow{2}{*}{86.88} \\
                                                                              &       &      &      &      &                                \\ \hline
\multirow{2}{*}{\begin{tabular}[c]{@{}c@{}}Ours \\ DinoV2\end{tabular}}       & Fit   & 85.4 & 98.3 & 91.5 & \multirow{2}{*}{\textbf{90.6}} \\ \cline{2-5}
                                                                              & Unfit & 96.9 & 91.9 & 89.5 &                                \\ \hline
\end{tabular}%
}
\end{table}

\section{Conclusion}
\label{sec:conclusion}

This research topic continues with the effort to raise new biometric applications and develop new complementary algorithms that may expand biometrics modalities' use cases.  This application can help to expand the knowledge of biometrics and capture devices in other areas, such as health, logistics, and mining applications, to prevent accidents.

This work was one of the first to propose and investigate the use of foundation models for the task of Fitness for Duty. We propose adapting two foundation models to this specific task under various levels of data availability.

Our experiments on multiple foundation models, training datasets, and a wide range of methods benchmarks led to interesting conclusions, which are summarised as follows:

a)~The FFD prediction based on NIR iris images is a challenging problem to solve, even for the most advanced foundation models.

b)~DinoV2 outperforms OpenClip in FFD prediction when using image data only, without any text input for OpenClip. However, the results from both models are competitive, even under the disadvantage of training from scratch using image data.

c)~Features extracted by DinoV2, utilising a self-supervised approach, are more effective than those derived from ImageNet and convolutional neural networks (CNNs).

d)~The drug detection in a LOO protocol obtained the lower error rate and appears to be the most confident class to be detected. Conversely, the alcohol class is the most difficult to classify.

In future work, we will explore applying Low-Rank Adaptation \cite{lora} of Large Language Models (LoRa) techniques or using a foundation model trained on face images, such as Facial Representation Learning to look at eye information in embeddings. 

\section*{Acknowledgements}
This research work has been partially funded by the European Union (EU) under G.A. no. CarMen (101168325) and the German Federal Ministry of Education and Research, and the Hessian Ministry of Higher Education, Research, Science and the Arts within their joint support of the National Research Center for Applied Cybersecurity ATHENE.

{\small
\bibliographystyle{ieee}
\bibliography{IJCB2025/ijcb}

\begin{thebibliography}{10}\itemsep=-1pt

\bibitem{Arora}
S.~S. Arora, M.~Vatsa, R.~Singh, and A.~Jain.
\newblock Iris recognition under alcohol influence: A preliminary study.
\newblock In {\em 2012 5th IAPR International Conference on Biometrics (ICB)}, pages 336--341, 2012.

\bibitem{FM-survey}
M.~Awais, M.~Naseer, S.~Khan, R.~M. Anwer, H.~Cholakkal, M.~Shah, M.-H. Yang, and F.~S. Khan.
\newblock Foundation models defining a new era in vision: a survey and outlook.
\newblock {\em IEEE Transactions on Pattern Analysis and Machine Intelligence}, pages 1--20, 2025.

\bibitem{CAUSA2024122808}
L.~Causa, J.~E. Tapia, A.~Valenzuela, D.~Benalcazar, E.~L. Droguett, and C.~Busch.
\newblock Analysis of behavioural curves to classify iris images under the influence of alcohol, drugs, and sleepiness conditions.
\newblock {\em Expert Systems with Applications}, 242:122808, 2024.

\bibitem{chen2023vlp}
F.-L. Chen, D.-Z. Zhang, M.-L. Han, X.-Y. Chen, J.~Shi, S.~Xu, and B.~Xu.
\newblock Vlp: A survey on vision-language pre-training.
\newblock {\em Machine Intelligence Research}, 20(1):38--56, 2023.

\bibitem{CHETTAOUI}
T.~Chettaoui, N.~Damer, and F.~Boutros.
\newblock Froundation: Are foundation models ready for face recognition?, 2024.

\bibitem{Czajka2015}
A.~Czajka.
\newblock Pupil dynamics for iris liveness detection.
\newblock {\em {IEEE} Trans. Information Forensics and Security}, 10(4):726--735, 2015.

\bibitem{imagenet}
J.~Deng, W.~Dong, R.~Socher, L.-J. Li, K.~Li, and L.~Fei-Fei.
\newblock Imagenet: A large-scale hierarchical image database.
\newblock In {\em 2009 IEEE conference on computer vision and pattern recognition}, pages 248--255. Ieee, 2009.

\bibitem{iris_chatgpt}
P.~Farmanifard and A.~Ross.
\newblock Chatgpt meets iris biometrics.
\newblock In {\em 2024 IEEE Internatl. Joint Conference on Biometrics (IJCB)}, pages 1--10, 2024.

\bibitem{iris-sam}
P.~Farmanifard and A.~Ross.
\newblock Iris-{SAM}: Iris segmentation using a foundation model, 2024.

\bibitem{FranzenBuysseDahlEtAl2009}
P.~L. Franzen, D.~J. Buysse, R.~E. Dahl, W.~Thompson, and G.~J. Siegle.
\newblock Sleep deprivation alters pupillary reactivity to emotional stimuli in healthy young adults.
\newblock {\em Biological Psychology}, 80(3):300 -- 305, 2009.

\bibitem{resnet}
K.~He, X.~Zhang, S.~Ren, and J.~Sun.
\newblock Deep residual learning for image recognition.
\newblock In {\em Proceedings of the IEEE conference on computer vision and pattern recognition}, pages 770--778, 2016.

\bibitem{Gelu}
D.~Hendrycks and K.~Gimpel.
\newblock Gaussian error linear units (gelus), 2023.

\bibitem{HollingsworthBowyerFlynn2009}
K.~Hollingsworth, K.~W. Bowyer, and P.~J. Flynn.
\newblock Pupil dilation degrades iris biometric performance.
\newblock {\em Comput. Vis. Image Underst.}, 113(1):150--157, July 2009.

\bibitem{mbv3}
A.~Howard, M.~Sandler, Chen, et~al.
\newblock Searching for mobilenetv3.
\newblock In {\em 2019 IEEE/CVF Internatl. Conference on Computer Vision (ICCV)}, pages 1314--1324, 2019.

\bibitem{DenseNet}
G.~Huang, Z.~Liu, L.~Van Der~Maaten, and K.~Q. Weinberger.
\newblock Densely connected convolutional networks.
\newblock In {\em 2017 IEEE Conference on Computer Vision and Pattern Recognition (CVPR)}, pages 2261--2269, 2017.

\bibitem{swin}
Z.~Liu, H.~Hu, Y.~Lin, Z.~Yao, Z.~Xie, Y.~Wei, J.~Ning, Y.~Cao, Z.~Zhang, L.~Dong, F.~Wei, and B.~Guo.
\newblock Swin transformer v2: Scaling up capacity and resolution.
\newblock In {\em 2022 IEEE/CVF Conference on Computer Vision and Pattern Recognition (CVPR)}, pages 11999--12009, 2022.

\bibitem{NIDA2021}
NIDA.
\newblock Is drug addiction treatment worth its cost?, June 2020.

\bibitem{dinov2}
M.~Oquab, T.~Darcet, T.~Moutakanni, H.~V. Vo, M.~Szafraniec, et~al.
\newblock {DINO}v2: Learning robust visual features without supervision.
\newblock {\em Transactions on Machine Learning Research}, 2024.
\newblock Featured Certification.

\bibitem{arc2face}
F.~Paraperas~Papantoniou, A.~Lattas, S.~Moschoglou, J.~Deng, B.~Kainz, and S.~Zafeiriou.
\newblock {Arc2Face}: A foundation model for {ID}-consistent human faces.
\newblock In {\em Proceedings of the European Conference on Computer Vision (ECCV)}, 2024.

\bibitem{Pinheiro}
H.~M. Pinheiro, R.~M. da~Costa, E.~N.~R. Camilo, A.~da~Silva~Soares, R.~Salvini, G.~T. Laureano, F.~A. Soares, and G.~Hua.
\newblock A new approach to detect use of alcohol through iris videos using computer vision.
\newblock In V.~Murino and E.~Puppo, editors, {\em Image Analysis and Processing --- ICIAP 2015}, pages 598--608, Cham, 2015. Springer International Publishing.

\bibitem{Radford2021LearningTV}
A.~Radford, J.~W. Kim, C.~Hallacy, A.~Ramesh, G.~Goh, S.~Agarwal, G.~Sastry, A.~Askell, P.~Mishkin, J.~Clark, G.~Krueger, and I.~Sutskever.
\newblock Learning transferable visual models from natural language supervision.
\newblock In {\em ICML}, 2021.

\bibitem{efficientnet}
M.~Tan and Q.~Le.
\newblock {E}fficient{N}et: Rethinking model scaling for convolutional neural networks.
\newblock In K.~Chaudhuri and R.~Salakhutdinov, editors, {\em Proceedings of the 36th Internatl. Conference on Machine Learning}, volume~97 of {\em Proceedings of Machine Learning Research}, pages 6105--6114. PMLR, 09--15 Jun 2019.

\bibitem{tapia2022alcohol}
J.~Tapia, E.~L. Droguett, and C.~Busch.
\newblock Alcohol consumption detection from periocular nir images using capsule network.
\newblock In {\em 26th Intl. Conf. on Pattern Recognition (ICPR)}, pages 959--966. IEEE, 2022.

\bibitem{TAPIA2025126511}
J.~E. Tapia, D.~Benalcazar, A.~Valenzuela, L.~Causa, E.~L. Droguett, and C.~Busch.
\newblock Classification of alcohol, drugs and sleepiness condition using periocular iris images to evaluate fitness for duty.
\newblock {\em Expert Systems with Applications}, 270:126511, 2025.

\bibitem{Tomeo-ReyesRossChandran2016}
I.~Tomeo-Reyes, A.~Ross, and V.~Chandran.
\newblock Investigating the impact of drug induced pupil dilation on automated iris recognition.
\newblock In {\em IEEE 8th Intl. Conf. on Biometrics Theory, Applications and Systems (BTAS)}, pages 1--8, Sept 2016.

\bibitem{lora}
B.~P. Veasey and A.~A. Amini.
\newblock Low-rank adaptation of pre-trained large vision models for improved lung nodule malignancy classification.
\newblock {\em IEEE Open Journal of Engineering in Medicine and Biology}, pages 1--9, 2025.

\bibitem{Zurita}
P.~C. Zurita, D.~P. Benalcazar, and J.~E. Tapia.
\newblock Fitness-for-duty classification using temporal sequences of iris periocular images.
\newblock In {\em 2023 11th International Workshop on Biometrics and Forensics (IWBF)}, pages 1--6, 2023.

\end{thebibliography}
}

\end{document}